# @ve: A Chatbot for Latin

Oliver Bendel[1] and Karim N'diaye[1]

[1] School of Business FHNW, 5210 Windisch, Switzerland
`oliver.bendel@fhnw.ch`

**Abstract.** Dead, extinct, and endangered languages have been preserved primarily through audio conservation and the collection and digitization of scripts and have been promoted through targeted language acquisition efforts. Another possibility would be to build conversational agents that can master these languages. This would provide an artificial, active conversational partner which has knowledge of the vocabulary and grammar, and one learns with it in a different way. The chatbot @ve, with which one can communicate in Latin, was developed in 2022/2023 based on GPT-3.0. It was additionally equipped with a manually created knowledge base. After conceptual groundwork, this paper presents the preparation and implementation of the project. In addition, it summarizes the test that a Latin expert conducted with the chatbot. A critical discussion elaborates advantages and disadvantages. @ve could be a new tool for teaching Latin in a memorable and entertaining way through dialogue. However, the present implementation is still too prone to glitches for stand-alone use – i.e., without the accompaniment of a teacher. The use of GPT-4 could be a solution as well as the extension of the knowledge base. In conclusion, it can be argued that conversational agents are an innovative approach to promoting and preserving languages.

## 1   Introduction

Human languages are cultural tools and cultural assets (Bendel 2023a). It is usually our concern to promote and preserve them, whether for functional or aesthetic reasons. Many languages are doing well – they are in demand and used on a large scale. They exist in spoken and, more recently in human history, in written form. English has had a unique success story, dominating vast areas of the world – particularly, the USA, Canada, Australia, and the UK – and of science. Chinese and Indian variants are of great importance simply because of their population size and the economic power of these countries (Eberhard et al. 2023).

Other languages are doing less well. They are threatened or endangered – or even completely extinct. They are neither in demand nor used on a large scale. Nevertheless, they can be very important for individuals and groups and not only from functional and aesthetic perspectives. Language is also home, is containment and demarcation to persons and groups: it creates identity. The loss of a language can have seri-ous consequences. With it, a cultural tool is lost as well as a cultural asset. In the worst case, an entire culture is lost. According to the linguistic anthology "Ethnologue", which regularly compiles an index of the most and least spoken languages, over 42% of the world's languages are endangered (Eberhard et al. 2023). According to the UNESCO Atlas of Endangered Languages, a distinction is made between "safe", "vulnerable", "definitely endangered", "severely endangered", "critically endangered", and "extinct" (Moseley and Alexandre 2010). Latin represents a special case. It is a dead language because there is no one who speaks it as a native language.

The main author of this paper has been developing conversational agents in the context of machine ethics and social robotics at his university since 2012 (Anderson and Anderson 2011; Bendel 2018, 2019, 2021). This has involved giving them moral rules or letting them show empathy and emotion to help humans. In 2022, another focus was added. The lead author and his rotating teams have been trying to use chatbots and voice assistants to rescue and promote endangered and dead languages. The idea is that schools and universities, as well as the general public, will be provided with such dialog systems. The users – i.e., the pupils and students – should be able to text or speak with them and thus learn a language or stay in practice. One could see the project in the context of "AI for Good" or "AI for the Good" (Vieweg 2021).

The first project in this series – which began in mid-2022 and ended in early 2023 – was devoted to Latin. Originally, it was supposed to be a chatbot for Rhaeto-Romanic, but there were too many technical difficulties at that time. Latin was an interesting alternative. It is still taught at schools and universities and is required in some places for admission to courses such as philosophy. Comics such as the Asterix booklet "Gallus" (2020) are a regular bestseller. Because of its popularity and the mass of texts, Latin seemed particularly suitable for an entry into chatbot development in this context. If one was successful here, one could also devote oneself to other languages and idioms. The chatbot developed was named @ve after the Latin word for "Greetings! Farewell!".

The timing for such a development was ideal. The lead author had been working on GPT-2 since 2019 and had included a colleague's contribution on the use of the language model in social robots in his edited volume (Bendel 2020). GPT-3 had already been published and was known for its increased effectiveness. At the end of the project, ChatGPT was to be made available to a world audience, drawing their attention to language models. Thus, large language models (LLM) were available that would most likely be able to cover the requirements of such a chatbot.

This paper starts with the basics of Latin and explains basic terms like "chatbots", "GPT-3", and "ChatGPT". In each case, it connects these with the given context. Then it presents the preparation and implementation of the @ve project. In addition, it includes the test of the chatbot by an expert of Latin. A critical discussion of the results follows. At the end there is a summary with an outlook.

## 2 Conceptual Basics

### 2.1 The Development and Importance of Latin

Latin is an Indo-European language (Clackson and Horrocks 2011). It was originally spoken by the Latins, the inhabitants of Latium with Rome as its center. Early Latin began to be spoken from the 7th or 6th century BCE, followed by Old Latin from the 3rd century BCE and Classical Latin – with prominent representatives such as Cicero and Ovid – from the first century BCE. As the official language of the Roman Empire, Latin became the lingua franca across the western Mediterranean.

Because of its importance for the linguistic and cultural development of Europe and because of the view that knowledge of the grammar facilitates understanding of the grammar of other languages, Latin is taught in grammar schools and universities, especially in Germany, Austria, and Switzerland. Moreover, it is taught in Italy at the grammar school level (Liceo classico and Liceo scientifico), i.e., in the country of origin of Latin (Oniga 2021). For some courses of study, such as philosophy, a profound knowledge of Latin or even the intermediate Latin certificate is required in some places. However, more and more universities are waiving this requirement. In the United Kingdom, Latin is already taught at the primary level in some private schools (Clackson and Horrocks 2011).

In the classroom, Latin is mainly translated into the standard language. The reverse is also not uncommon. The students or the teachers and lecturers also read the texts aloud. Rarely do the students have to create their own texts and try them out in conversation inside and outside the institutions. What is missing is something vital for any living language: application in everyday situations. Through such a situation, however, a language is much better learned and retained. And that is also where it fulfills its actual purpose, which is communication.

### 2.2 Chatbots as Conversational Agents

Chatbots are dialog systems with natural language capabilities of a textual or auditory nature (Kohne et al. 2020; Bendel 2017). They are used, often in combination with static or animated avatars, on websites or in instant messaging systems, where they explain and advertise the products and services of their operators or take care of the concerns of interested parties and customers – or simply serve for amusement and reflection. Sometimes the term chatbots is so broadly defined that it also includes voice assistants with voice input and output. Chatbots can be rule-based and access a knowledge base with statements and answers or be located in the field of artificial intelligence (AI) (Chowdhary 2020) and use methods such as machine learning. These can also be termed as conversational agents or conversational AI (McTear 2020; Bendel 2023b).

Chatbots are often available in English and Chinese. This covers numerous countries and language communities. German, French, and Spanish chatbots are also available. Users in Europe particularly appreciate being able to communicate in the language of their country. Spanish and French, on the other hand, are also world languages. Since chatbots are primarily used for business (such as addressing customers and providing information) and entertainment, endangered or extinct languages have not been in view so far. Development of a classical rule-based chatbot would also have been too time-consuming and cost-intensive. Machine learning and especially language models offer more powerful and at the same time simpler possibilities here.

## 2.3   The Language Model GPT-3

GPT-3 was introduced on June 11, 2020, and is the third in a series of language models capable of generating intelligible and meaningful text (Dale 2021). It is based on the Transformer architecture and has been trained by generative pre-training with a very large amount of data (Gillioz et al. 2020). GPT-3 differs from other models mainly in its size and performance capabilities. With 175 billion parameters – also called neurons – it was the largest language model to date until the release of GPT-4 and comparable models. The transformer architecture in combination with the amount of pre-trained data enables GPT-3 and GPT-4 to solve very complex tasks in the field of Natural Language Processing (NLP), which were previously impossible or difficult for smaller models.

The large amount of data also includes Latin texts. Apparently, OpenAI has scanned entire books – in principle, classics such as Ovid's "Metamorphoses" or Julius Caesar's "De bello Gallico" are available. Social media could not be relied upon in this case, nor chat conversations – Latin is hardly actively used except by the church and beyond the educational sphere. The language model allows the user to train it with further data, such as teaching materials for Latin. It must be ensured that these are copyright-free or may be used for this purpose.

## 2.4   The GPT-based Application ChatGPT

"ChatGPT" stands for "Chat" and "Generative Pre-trained Transformer" (Bendel 2023c). It is a chatbot (or content production system) from OpenAI based on the same company's GPT-3.5 language model. The training data comes from forums, articles, books, and spoken language, among other sources. A special form of machine learning is used, namely Reinforcement Learning from Human Feedback (RLHF). This involves people (mainly low-wage workers in emerging countries) who find certain answers to be good and correct. Their feedback is used to train a reward system, which in turn trains the chatbot. The end user can also evaluate the quality of answers.

After its release as a prototype in November 2022, ChatGPT quickly gained high attention and popularity. This was helped by the fact that the system could be tried out by millions of users. This meant that, after translation programs such as DeepL, artificial intelligence dealing with language had finally reached the general public. The strengths of ChatGPT quickly became apparent, e.g., in the creation of texts or in programming, but also its weaknesses, especially with regard to facts – for example, people are simply attributed professions and memberships or sources are freely invented. Another limitation was that it only processed data up to the year 2021.

ChatGPT has mastered Latin to some extent. One can elicit Latin sentences from it – which it explains again and again in English – and ask the chatbot about history and grammar. So, one could use it to some extent in the classroom and thus also to revital-ize Latin. However, the chatbot was not yet available at the beginning of the project. In addition, it is probably always important not to rely solely on a commercial version – its use can be problematic for schools, in particular for data protection reasons, and it can change its business model at any time – and to provide for authoritative answers via a specially created knowledge base or knowledge database.

## 3  Preparation of the @ve Project

The @ve project was initiated in mid-2022 at the School of Business FHNW by the main author. He also acted as the client, with the corresponding rights (use and further development of the prototype) and duties (provision of a budget). He was able to collaborate with the co-author for the implementation. Until then, the developer had no knowledge of Latin and, above all, the motivation to take a closer look at GPT-3 and possible applications.

The chatbot was named @ve at the suggestion of the main author, after the Latin word for "Greetings! Farewell!". The name is also reminiscent of Ava, the artificial intelligence equipped robot from the movie "Ex Machina" (2015). The main author was available to provide expertise to the co-author throughout the project. Technical details were also discussed with an additional supervisor, Bradley Richards.

The project lasted half a year, from August 2022 to January 2023, with the final presentation and acceptance scheduled for this last month. There was a small budget of 500 Swiss francs, which is equivalent to about 500 US dollars, provided by the client.

### 3.1  Requirements Analysis

The requirements that form the basis for the implementation of the objective were predefined in the first phase of the project and refined through discussions with the client – the main author (N'diaye 2023). They are listed below.

– Functionality is crucial when evaluating the technological basis of the chatbot. It must be able to communicate primarily in Latin. This means that it must be able to form complete, understandable sentences in this language and give suitable answers to questions in it.
– If the chatbot could also communicate in German or English, this would be ideal for introducing it to people who do not yet have a strong knowledge of Latin. They can ask something about the grammar of Latin in these languages. A deeper knowledge about the history, grammar, and spelling would also be desirable of the chatbot.
– Ease of use is another factor to consider. The chatbot should be simple to use, the chat window should be easily visible, and its responses should be easy to read and understand. If it is difficult to use, users may not be able or willing to take advantage of all the features and benefits.
– Performance is also relevant. The chatbot should function smoothly and efficiently to generate quick responses. Insufficient response times lead to the user be-coming frustrated and abandoning the dialog. In addition, there is not unlimited time available in class or during a preparation phase.
- The longevity of the underlying technology – in this case GPT-3 or the related framework – is also important. It would be disadvantageous if the chatbot became obsolete shortly after release, especially since schools and universities often have limited options for use and change.
- The power of the underlying technology also plays a role, because the more interesting, varied, and in-depth the conversations with the chatbot are, the more likely it is to be used recurrently. This also calls for the open domain approach, according to which the chatbot can master a wide range of topics.
- Last but not least, the relevant programming interfaces (API) must be given so that the chatbot can be connected to a website or to a messenger service. The prototype should at least be connected to a website and thus be available to a tester and an interested audience.

The listed requirements were not prioritized, as all of them had to be met within the project if possible.

### 3.2 Utility Analysis

To ensure the best possible basis for @ve, a utility analysis was carried out comparing three solution approaches, namely the use of IBM Watson, the manual creation of a chatbot with GPT-3 as the basis and Python as the programming language, and the implementation of a Quickchat chatbot (N'diaye 2023). Other candidates had previously been tested for compatibility with the above requirements and then rejected.

**IBM Watson.** Relatively early in the project, a personal contact with an employee of IBM Switzerland was used to check the possibilities for the use of an IBM Assistant chatbot primarily specialized in customer support (IBM w.d.). By providing a test access for the IBM cloud platform, the presented requirements could be tested.

However, IBM's Watson Assistant did not meet the requirements for this project because of its strongly rule-based functionality. This has to do with the fact that it is often used in customer support, where it functions as a source of information for medical concerns, for example, and needs a high degree of reliability. In this infrastructure, it would have been highly laborious to build up its skills in Latin.

**GPT-3 and Python.** Initial test trials via a GPT-3 test environment were very promising. It quickly became clear that the language model has the ability to conduct a conversation in Latin. This would drastically reduce the effort, while an additional knowledge base would still be useful and feasible.

Problems were found in the calculation of the effort that results from the manual development of a chatbot including knowledge database and integration into a website. There was a risk that the project goals could not be achieved within the timeframe.

**Quickchat.** Quickchat, a startup specializing in AI-powered customer support chatbots, offers an interesting approach (www.quickchat.ai). With the help of GPT-3 and using a knowledge database, a chatbot can be developed that can transmit information from the stored database and the already trained knowledge of GPT-3 remarkably well.

Tests with Quickchat were very promising right from the start and the requirements could be met to a large extent, which is also reflected in the final evaluation of the utility analysis (N'diaye 2023). The costs could be covered by the available budget over the required term and were therefore manageable.

After conducting the utility analysis, the decision was made in favor of Quickchat. This was approved by the client, the developer, and the additional support person. Subsequently, it also proved to be an advantage to be able to submit support requests, which were answered quickly.

## 4      Implementation of @ve

This chapter discusses the specific implementation process and explains how @ve was integrated into the website. It was decided early in the project that the chatbot should be made available on the website www.ave-bot.ch (N'diaye 2023). The main reason for this was that it would provide the most accessible access and independence from messenger services. With this public provision, testers and interested parties could be served. In addition, however, at least one messenger connection was to be implemented.

### 4.1      GPT-3 as the Basis for @ve

Through the utility analysis, Quickchat was determined to implement @ve. The company offers a modular system for chatbots. It is based on GPT-3, which in the case of @ve is GPT-3.0, while ChatGPT, which was released later, is based on an improved version, GPT-3.5.

The chatbots created with the help of Quickchat are primarily intended for use in customer support, where they provide information about processes and products to customers with the help of a knowledge database. Therefore, a knowledge database was available by default. This was converted for @ve so that it consists purely of information about the Latin language, such as its history and grammar.

Quickchat's technical support was helpful in enabling bilingualism, which is by no means the norm. This meant that @ve could write not only in Latin, but also in German, which is a great advantage for German-speaking users who are still in the learning process, because it means that information about Latin can be requested in their mother tongue. One could also use @ve in English.

### 4.2      Design of the Chat Window

The chatbot itself was realized in conventional form (N'diaye 2023). The chat window with rounded corners constantly shows a circle with an avatar, an alienated image of a young, modern looking woman (see Figure 1). This was taken from the im-age platform Pixabay, where free images are available. An image of a woman was obvious because the name of the chatbot appears feminine.

Inputs and outputs are shown linearly, with the most recent at the bottom. @ve's statements and responses are left-aligned and underlaid with gray shading, while the user's statements and questions are right-aligned and underlaid with blue shading. The chat window always remains at the bottom right of the web page.

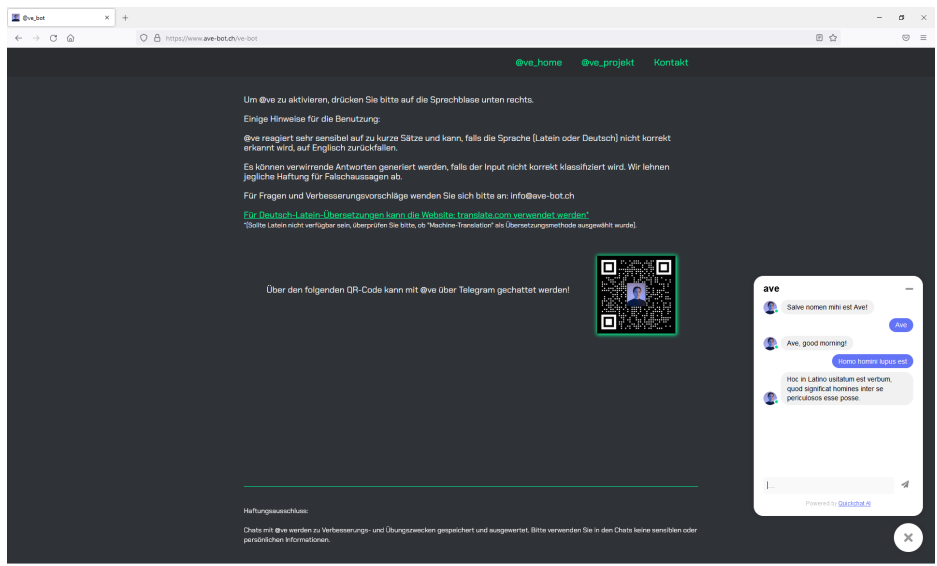

Fig. 1: The web page with the information and the chat window with @ve

### 4.3   Integration of @ve into the Website

The chatbot was published on the website www.ave-bot.ch and thus made accessible to an interested audience (N'diaye 2023). The website as the main access point also made it possible to achieve independence from messenger services. Any user can access @ve through a browser, on a notebook, on a tablet, or on a smartphone. This also eliminated the need for a native app, which is often associated with privacy issues.

The connection to the website www.ave-bot.ch was done via a JavaScript provided by Quickchat, which was placed in the head tag of the desired page – @ve_bot. The head tag is an HTML element that contains metadata about a web page, such as the page title, scripts, and stylesheets. By placing it exclusively on the @ve_bot page, it was possible to ensure that a conversation could not be started on the home page. The main reason for this is that it allows users to see the handling information.

Clicking on "Click here to chat with @ve!" on the home page brings up the handling information page. The most important hint comes right at the beginning. It says,

"To activate @ve, please press the speech bubble at the bottom right." (Translation by the authors) At the bottom right, there is a picture of @ve with the speech bubble. It goes on to say, "@ve is very sensitive to sentences that are too short and may fall back to English if the language (Latin or German) is not recognized correctly. Confusing responses may be generated if the input is not classified correctly. We disclaim any liability for misstatements." (Translation by the authors)

In addition to the handling information page, there is a page with information on the motivation and background of @ve. Regarding the motivation, it says: "Preserving dead, extinct, and endangered languages is an important concern to us. @ve is a first attempt in this direction. The chatbot has already learned Latin quite well with the help of GPT-3 and, in the best case, ensures that it is mastered by more people than before." (Translation by the authors) In addition, there is a page with contact options. Besides photos of the main project contributors – main author and co-author – there is an email address.

### 4.4 Telegram Connection

The connection to Telegram was done as an optional step (N'diaye 2023). This was to show that @ve works in other environments as well. Moreover, messengers are a common and indispensable application for many people, and one can just about say that there has been a mass "move" of chatbots from websites to messengers since the turn of the millennium.

To integrate a Quickchat chatbot into the Messenger service, a "telegram bot token" must be generated for it via the administration interface. The next step involves launching Telegram's interactive bot creation solution, "Bot-Father". A login link can then be created, which in the next step can be transformed into a QR code by a provider such as qrcode-monkey.com and embedded into the website.

The QR code is prominently placed in the right part of the center. It has white dots on a black background instead of the usual black squares and rectangles to look futuristic. The three large squares in the corners – the so-called position markers – are also inverted. In the center, the colorful image of @ve has been placed. In principle, the QR code can be saved and shared as an image. In this way, it can spread on social media, for example.

## 5 Knowledge Base of @ve

In this chapter, the knowledge base of @ve is discussed (N'diaye 2023). This is a central element of the application and allows targeted extensions. The manually created knowledge base gives the chatbot a broader understanding of the history, grammar, and culture of the Latin language than is the case in the standard execution of GPT-3. This allows specific learner needs to be addressed. In addition, specific answers to specific questions can be forced.

### 5.1 Structure of the Knowledge Base

The information that was collected to populate the knowledge base was mostly taken from Wikipedia. It was collected there and then manually formatted. In a final step, the information was uploaded to the Quickchat platform, where @ve could be trained on it.

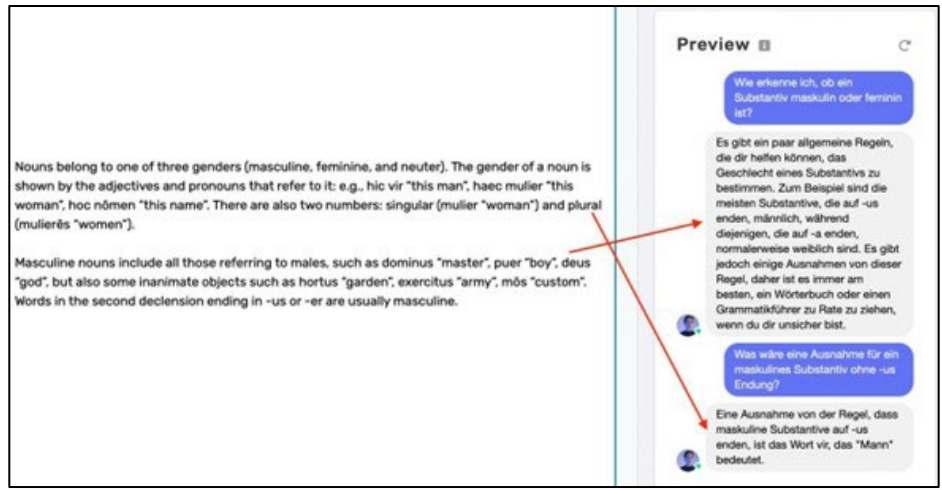

Fig. 2: Knowledge Base of @ve

The screenshot in Figure 2 shows a section of @ve's knowledge base on Quickchat's administration interface. It is important to note that the chatbot processes information in English and consequently this is how the knowledge base was created. The screenshot also shows that @ve is accessing the stored information in the conversation on the right half of the screen.

### 5.2 Information Search

The knowledge base can only be populated by unformatted text corpora, each no longer than 140 characters (N'diaye 2023). It should be noted that the information contained in the text corpora must have a certain coherence, which means that, for example, a text cannot be about adjectives and about Cicero – the Roman politician, writer, and philosopher – at the same time.

The information search was facilitated by a Python program that allowed the unformatted text to be downloaded from Wikipedia pages and saved locally as TXT files. The program was developed by the co-author himself, also using GPT-3. It was not very complex, and its only purpose was to avoid having to manually copy the text from Wikipedia.

### 5.3 Information Processing

The greatest effort was put into formatting and breaking down the raw text files in a Word document, dividing them into units of no more than 140 characters each and with one piece of information per unit (N'diaye 2023). This was done manually.

The resulting text files enabled @ve to train the information in a targeted and clean manner. Preparing the texts for the knowledge base allows @ve to respond specifically to user questions. Consequently, preparing the data can create a better user experience.

### 5.4 Information Access

The knowledge base, as has become clear, has a dual function. In certain cases, @ve accesses entries directly and reproduces them verbatim or approximately verbatim. Thus, it behaves like a classic rule-based chatbot that does not use machine learning or deep learning.

In addition, @ve constantly increases its general knowledge, if one wants to speak of knowledge, by training with the contents of the knowledge base. This, in turn, can be continued by adding more data. The chatbot thus becomes a Latin teacher who is not omniscient, but still appears educated.

### 5.5 Creativity Setting

The creativity setting in the dashboard determines how often the AI says "I don't know" or tries to find a plausible answer based on the available knowledge. There are three levels in total, namely low ("outright saying the AI does not have sufficient information to answer"), normal ("balance saying 'I don't know' and improvising answers"), and high ("try and improvise an answer based on the knowledge base").

If the tester or user finds that the chatbot occasionally provides incorrect answers, creativity can be reduced by the developer. On the other hand, if the AI answers "I don't know" too often, creativity can be increased. The project team decided on the highest level, which, according to the definition of Quickchat, is also allowed to improvise.

## 6 Expert Test of @ve

@ve was tested by the main author and especially by the co-author (N'diaye 2023). The main issue was whether a conversation in Latin is basically possible and whether questions about grammar and history are answered correctly. The main author had some knowledge of Latin, the co-author only little – but he used different translation programs for the test, so that he could understand the quality of the answers in Latin. However, a test with an expert of the language seemed necessary.

In order to answer the question about the prerequisites, the advantages and the disadvantages of an insertion, an expert for the Latin language was consulted. This expert was Ms. Z., who holds a Master of Arts in History and Latin Linguistics and Literature as a relevant qualification for the evaluation and does not wish to be named.

It was determined that Ms. Z. would conduct a conversation with @ve independently as much as possible without bias. She should only ask questions in case of technical difficulties. Afterwards, she would describe her collected impressions and make a short analysis of the errors @ve encountered. In addition, she would list advantages and disadvantages.

In the course of the conversation, some weak points of @ve became apparent. At the beginning, for example, there were situations in which @ve switched to English because of a single word or a sentence from Ms. Z. that was too short. Furthermore, a predicate was completely omitted in an answer sentence. Sentence construction errors and other grammatical errors were repeatedly a problem, but there were also passages in which Ms. Z. and @ve were able to talk about stories by the author Ovid without errors.

Ms. Z. sees the advantages of a conversation with @ve in the fact that the users can formulate sentences themselves and thus they have to think about the forms exactly. In addition, @ve could be used well in a school context, especially since, in the expert's opinion, Latin lessons should be enriched with more active elements. Another positive point is that, according to empirical studies, communication with a counterpart (here only a virtual one) promotes the elaboration process, which means that learning content passes more easily into long-term memory in this way. The expert thus confirms the findings reproduced above.

The disadvantages, according to Ms. Z., are that @ve cannot be asked why certain forms are used. Another problem she sees is that @ve does not correct users, which can lead to their own mistakes being deepened without this being noticed. The grammatical errors and the errors in conversation – jumping into English – were noted by Ms. Z. and were definitely seen as a problem.

Following on from this test, it can be said that @ve has considerable potential, but mainly in a setting that allows the chat history to be monitored by a trained person, for example, in a school class where, after the interaction between the students and @ve is complete, the chat logs are evaluated and discussed by the teacher. Overall, Ms. Z was positively surprised by @ve: "The chatbot definitely provides good entertainment! It works well." (Translation by the authors)

## 7  Discussion of the Results

One of @ve's strengths is her manually created knowledge database. This provides the chatbot with specific knowledge about the language, such as history and grammar. Wikipedia as a source of information can certainly be faulted. The advantage was that extensive knowledge was available and the use was legally unobjectionable and free of charge. In addition, the encyclopedia is relatively reliable in such areas – which can be judged by a large number of users and are readily judged by academics.

From the conversations of the test subjects and the anonymous users, it can be concluded in principle that @ve is capable of conversing about a wide range of topics. It is also capable of providing specific information about Latin. At times, however, there has been abuse, such as when @ve was asked for Hitler's address (and then replied that she did not have it, but that she did have the phone number). Such abuse could easily be prevented, for example, by an exclusion list or specific training.

Testing in a school class or in a university classroom was not possible as part of the project but would be useful. It could be done by the lead author at any time in the future if a request comes in or the opportunity arises. The chatbot will continue to be hosted on the www.ave-bot.ch website – probably until the end of 2023 – and possibly beyond if funding can be found.

The results of @ve are basically promising, but the criticisms of Ms. Z. must not be neglected. @ve does not offer error correction, which is a problem if the chatbot is to be offered in a school context. It is important that, unless it is just a pastime, it is supervised by a qualified teacher who can provide help, advice, and correction suggestions to learners.

Another danger addressed by Ms. Z. is the reinforcement of one's own spelling and grammatical errors. By not pointing out errors to @ve users, these can be memorized and thus reinforced. However, it is also important to remember that additional teaching and learning materials are usually available to help identify errors. The teacher can point them out more.

In the project, it was not possible to call in several experts due to time constraints. However, the exchange with Ms. Z. alone was very profitable, and it is questionable whether more experts would have provided more insights at this stage. Ultimately, each conversation with a GPT-3-based chatbot is only a snapshot in time, as it can turn out differently in each case, even if the knowledge base catches this to some extent.

What is unique is that learning Latin is now possible with a new, active component. Until now, people mainly translated and discussed texts. Even creating one's own texts was not common. With @ve a communicative partner is available. One has to think about and enter one's own sentences and immediately gets an answer, to which one can react in turn. Besides the learning effect, the fun factor should be emphasized.

## 8  Summary and Outlook

The present contribution to the project on the chatbot @ve first established basics on Latin and explained basic terms such as "chatbots", "GPT-3", and "ChatGPT" and connected them with the present context. Then, it presented the preparation and implementation of the project, up to the integration in the website and in a messenger. In addition, the test of the chatbot by an expert of Latin was discussed. This was followed by a critical discussion of the results.

With @ve, for the first time, a chatbot based on GPT was created for Latin. This brought a dead language to life in that users can communicate with her. Teaching and learning materials on Latin usually do not have this interactive component. Thus, the language is practiced and learned in a very limited way. The project's findings could be applied to chatbots for endangered languages, for example.

This was exactly the next step. In February 2023, a second project was started that focuses on Rhaeto-Romanic, one of the four national languages of Switzerland. Thus, the original plan was resumed. There are radio and television programs on Rhaeto-Romanic and the radio station Radiotelevisiun Svizra Rumantscha (RTR). An artificial language for writing has been created, Rumantsch Grischun, which is used by the authorities and for subtitling films and broadcasts. Nevertheless, the language is considered endangered. With @llegra, one of the main idioms, Vallader, is to be strengthened. It is a GPT-4-based chatbot that also uses (but does not understand) spoken language, thanks to a specialized text-to-speech engine that is used in the project.

The present implementation of @ve is still too glitch prone for stand-alone use. One could use her together with a teacher or instruct the students accordingly. One could also develop @ve II, using GPT-4 and extending the knowledge base. In cooperation with publishers, further texts could be included in the knowledge database, such as primary literature by Cicero and Caesar or literature on the grammar of Latin. Accordingly, the project as a whole would have to be commercialized. @ve I, if one wants to call her that, has basically shown the potential for the preservation of dead and endangered languages.